\documentclass[conference]{IEEEtran}
\IEEEoverridecommandlockouts
\usepackage{cite}
\usepackage{verbatim}
\usepackage{enumitem}
\usepackage{amsmath,amssymb,amsfonts}
\usepackage{algorithmic}
\usepackage{graphicx}
\usepackage{textcomp}
\usepackage{xcolor}
\usepackage{hyperref}

\def\BibTeX{{\rm B\kern-.05em{\sc i\kern-.025em b}\kern-.08em
    T\kern-.1667em\lower.7ex\hbox{E}\kern-.125emX}}

\title{Lessons Learnt from Deploying ARI in Residential Care

\author{Sara Cooper$^{1}$, Francesco Ferro$^{1}$}
\thanks{*This work has been partially funded by SHAPES project,  which has received funding from the European  Union’s Horizon 2020 research and innovation  program under grant agreement no. 857159. }

\thanks{$^{1}$, 
{PAL Robotics, Barcelona, Spain {\tt \{sara.cooper, francesco.ferro\}@pal-robotics.com}}}

}

\begin{document}
\maketitle

\begin{abstract}

This paper describes the final prototype of an assistive robot used for increasing engagement of older adults in the context of SHAPES project. It then highlights lessons learned from hands-on training during the first phases of the pilots at Clinica Humana and Ca’n Granada residence in Mallorca (Spain). 

\end{abstract}


\section{Introduction}

As people age,  more support is needed to live independently, for everyday tasks like reminders to increase social engagement and reduce isolation.  Social robots~\cite{tapus2007socially, robinson2014role} are an alternative to digital assistants like Alexa or Google that have the potential to support older adult care, thanks to their capability of navigation and proactiveness. For this a co-design user-centered design of social robotic applications is needed. SHAPES Project \footnote{\url{https://cordis.europa.eu/project/id/857159}} is specifically focused on integrating a set of digital solutions to promote healthy living, and  PAL Robotics ARI robot \cite{cooper2020ari} has been used for one of the pilot deployments, specifically, in PT1-004~\cite{cooper2021social}.

Each SHAPES pilot consists of 5 phases~\cite{spargo2021shaping}: Phase 1 (Design), Phase 2 (mock-ups), Phase 3 (Prototype and hands-on training), Phase 4 (selected pre-pilot testing) and Phase 5 (final pilot). 

In Phase 2, mock-up phase, described in \cite{cooper2021social}, power points and possible activities that the robot could carry out were shown to care-givers and older adults. From there, the robot's front-end interfaces have been improved and almost all the defined features have been finalized. At last two use-case scenarios have been defined: 

\begin{itemize}
\item Robot is co-living with an older individual at their private home. In this context, the robot is aimed at providing entertainment games, reminders, agenda, videocall and alerts - too high temperature or falls detected that send an SMS to respective care-givers. 
\item Robot is co-living at sheltered apartments with individual users. In this context, ARI is in the hall performing tasks such as providing information about week activities, registering menus, providing information about the care-home, and so on. Here care-givers are in charge of filling the robot's content such as menu, events they can sign up for, and receive the responses as emails.
\end{itemize}

This paper focuses on the development and results of Phase 3, where the prototype has been completed and older adults and care-givers have managed to do hands-on training with the robot.
The next section will first describe the final prototype implementation (see Section II), followed by the hands-on training execution and results (see Section III). It then indicates future work for Phases 4 and 5 together with conclusions (see Section IV).

\section{Prototype development}

The robot is endowed with two behaviors, developed by means of Behavior Trees~\cite{iovino2020survey}\footnote{https://github.com/BehaviorTree/BehaviorTree.CPP}. The first behavior consists in users starting the interaction with the robot that is static at the reception or docking station; while in the second behavior the robot actively looks for a user in a proactive way to start the interaction. 

\begin{figure}[h]
    \centering
    \includegraphics[width=\columnwidth]{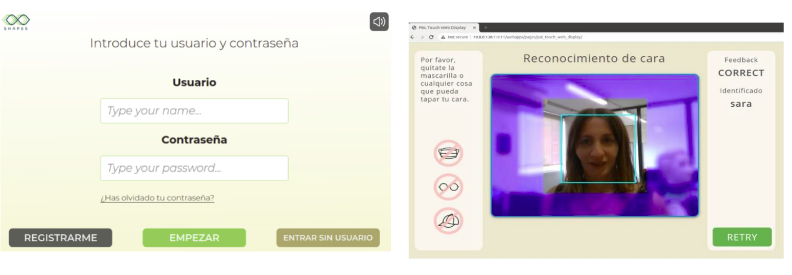}
    \caption{SHAPES login page and face recognizer}
    \label{fig:login}
\end{figure}

For the second behavior, the robot starts starts by looking down, staying inactive, until the user presses the touch-screen. At that point, the robot starts identifying the face using a Face Recognition solution from VICOMTECH \cite{goenetxea2021efficient,elordi2021optimal}. If this is not successful after 30 seconds, the robot prompts the user to introduce the SHAPES username and password (Figure~\ref{fig:login}) to log to ASAPA, i.e., SHAPES authentication platform.

This unique authentication from the project ensures user information is kept private and for the robot to customize its behavior to the user. 

Once logged in, the robot saves the user token, that is refreshed constantly unless the robot determines the interaction has finalized (through a user engagement module) or the user manually logs out. 

\begin{figure}[h]
    \centering
    \includegraphics[width=\columnwidth]{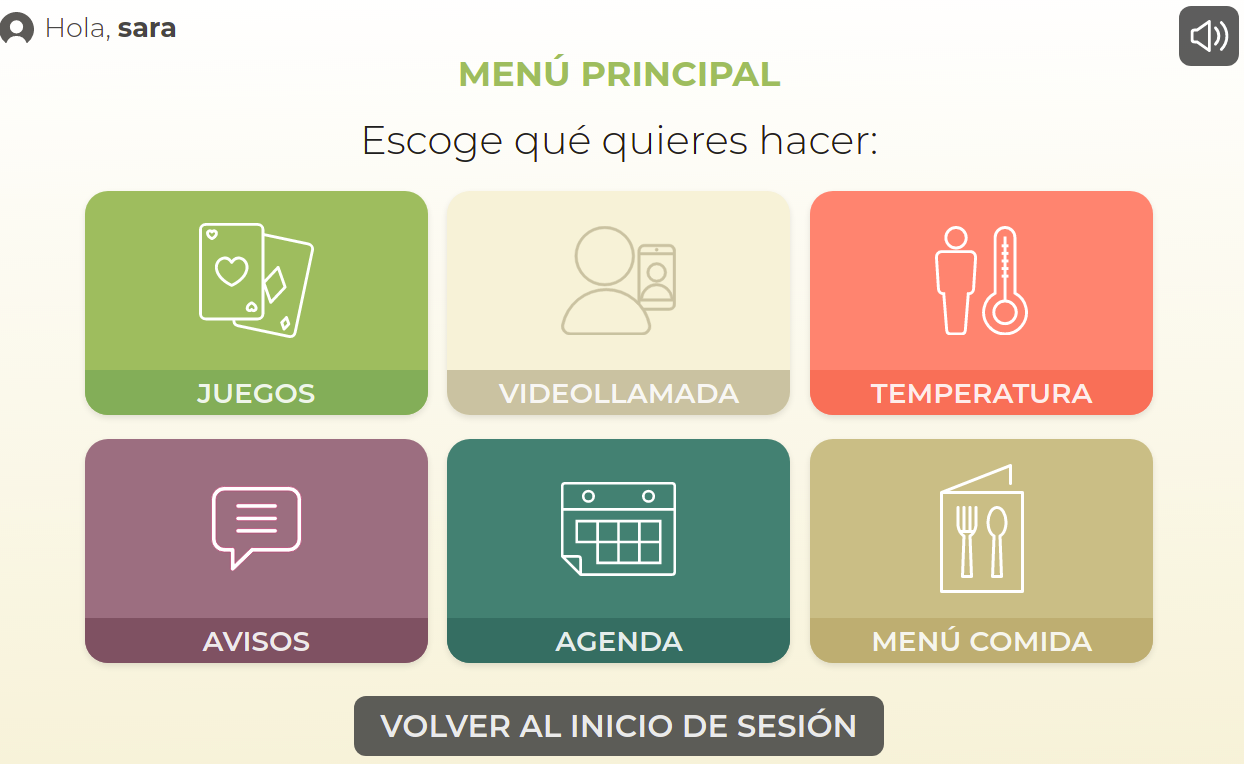}
    \caption{Pilot main menu displayed on ARI touch-screen.}
    \label{fig:apps}
\end{figure}

\begin{figure}[h]
    \centering
    \includegraphics[width=\columnwidth]{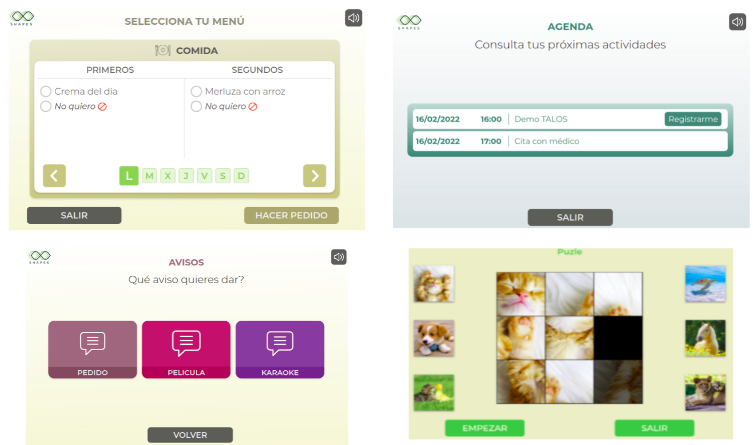}
    \caption{Improved front-end for Phase 3 including food menu (top-left), agenda (top right), alerts by email (bottom left), game (bottom right)}
    \label{fig:apps_options}
\end{figure}

From here, the user is shown the main menu (Figure~\ref{fig:apps}), which is slightly different for each of the two situations described in Section II. An example of the activities is shown in Figure~\ref{fig:apps_options}.

\begin{itemize}
    \item \textit{Visualizing the agenda of the day}: events can be introduced through a Google Excel sheet that the robot has access to, and are displayed as pop-ups when the user logs in or visualized as a list. 
    \item \textit{Filling in the weekly menu}: a request from Phase 2 mock-ups, as in care-homes so far they had to manually go to each user to ask their daily menu. Users can fill in the menu using the robot's touch-screen, and the robot then updates a Google Excel spreadsheet, before sending an email to the caregivers indicating who has filled in the menu.
    \item \textit{Sending alerts}: with option to send an email to the user’s caregiver, both for health-related information but also if they wish to sign up for care-home events like movies, karaoke, etc.
    \item \textit{Videocalling}: the robot integrates meet.jit.si, an online videocalling platform. At this stage, however, video and audio quality are still being improved
    \item Temperature monitoring: a COVID-19 solution request, where the robot uses its thermal camera of the head to measure temperature, and offers the option to call the respective caregiver if needed using Twillio \footnote{\url{https://www.twilio.com/}}
    \item \textit{Playing games}: entertainment games like puzzles, solitaire or finding matching pairs, where the robot offers feedback.
    \item \textit{Information about Can Granada}: addition requested from Phase 2 mock-ups, where the robot provides general information with touch-screen content and speech about the care-home facilities.
\end{itemize}

Together with the above, the robot integrates a voice-based assistant from VICOMTECH (Adilib platform)\cite{serras2019user}, so that all touch-inputs can also be said through voice. Soon the wake-up word will also be integrated to facilitate speech-based interaction. Moreover, robot behavior will be enhanced so it indicates when it is listening by touch-screen icon, LED color change and head direction.

\begin{figure}[h]
    \centering
    \includegraphics[width=\columnwidth]{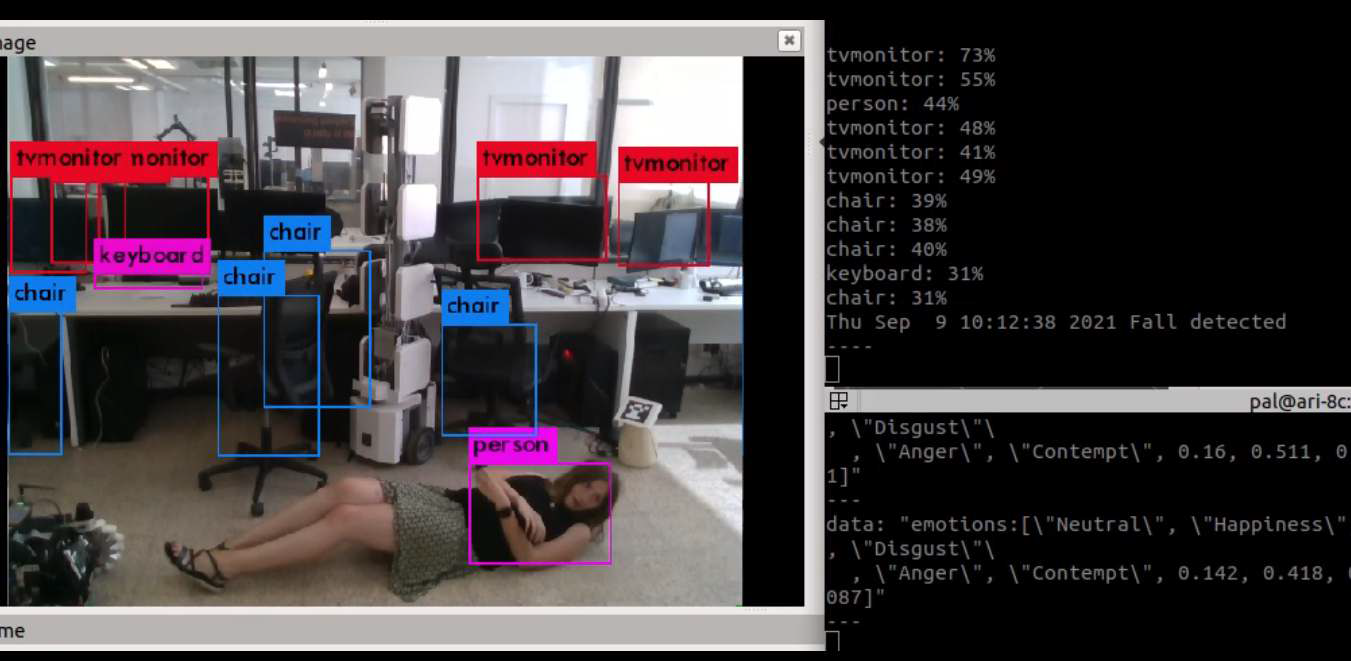}
    \caption{TREE Technology fall detector}
    \label{fig:falldetection}
\end{figure}

Furthermore it uses emotion recognition from TREE Technology \footnote{\url{https://www.treelogic.com/en/Computer_Vision.html}}and fall detection Figure~\ref{fig:falldetection}. 
Asides from the idle behavior, the robot has another behavior where it is on stand-by until it needs to deliver a reminder. At that point, it navigates through a pre-defined trajectory using Visual SLAM, as it uses people detection (TREE Technology) and face recognition to find the required user.

\section{Hands-on training at Clinica Humana (Mallorca)}

The goal of Phase 3 hands-on training of SHAPES Pilots is to collect feedback (user experience) from end-users -older adults and caregivers- by giving them the option to try the robot before it is deployed in the use case PT1-004. 

Phase 3 hands-on experiments were conducted with 5 target users of ARI (i.e., $\geq$ 65 years’ old). For the tests, the robot was first taken to the Can Granada residence and the different functionalities mentioned were demonstrated step by step together with a presentation. 

After each point of the demonstration, the participants were encouraged to use ARI following the same process, with the presenter still present and available to be asked questions and troubleshoot any issues.

In general ARI raised positive interest from the end-users and the demos were well-liked. On the other hand, ARI is usually seen as too big. We oversee that an accommodation time will be needed, probably in close contact with a reference guide who will instruct and support the older person the first days. Some people did not feel comfortable with the eye movement, and would prefer static eyes or even shutting them down - adjustable eyes will be integrated for the next phase. 
An aspect that was mentioned by the caregivers was the need to have pop-ups in order to confirm the action, e.g., before logging out, sending the weekly menu, sending an alert to the caregiver. It was also indicated that the robot should have a more adaptive interface to increase its accessibility: different font sizes for the touch-screen, adjustable volume, or a mechanism to compensate for the height difference between people sitting down or in wheelchairs and the robot’s touch-screen - even if for some using a tablet pencil was enough. Some solutions that have been integrated to counter this include icons to adjust volume, or an additional Android tablet at the back.  

Furthermore, the robot should customize more to the user. For example, they noted that if a user has already played solitaire a few times, the robot should remember it in order to skip the initial instructions, and should also remember the configuration chosen by the user (volume, font size).  

\begin{figure}[h]
    \centering
    \includegraphics[width=\columnwidth]{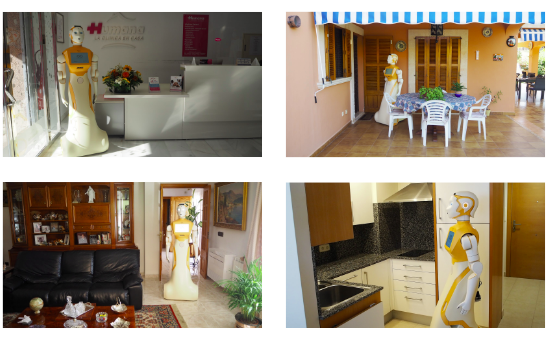}
    \caption{ARI at Clinica Humana and Ca'n Granada care-home}
    \label{fig:navigation}
\end{figure}

On a more technical level, fall detection was considered to be difficult to evaluate, mainly due to navigation restrictions. Older adults' individual homes most often had carpets, narrow corridors, that required to restrict the pre-defined trajectory the robot could make. Figure \ref{fig:navigation} shows some examples of challenges produced through navigation.

It was requested to use the robot outside at the terrace, but steps or risk of wet areas prevented it. Moreover, older adults appeared to be reluctant to move furniture around.  For this reason the proposed trajectory reduces the likelihood of the robot finding the user, even more so to detect if they have fallen down (e.g. more likely in the toilet, or room, areas where the robot will not enter). For this reason fall detection will not be used in the next phases, and the robot will only navigate in individual homes when it needs to provide a reminder. 

When it comes to the registration and authentication procedure to SHAPES platform, participants found it easy to do the log-in through credentials and then face recognition. Two points of improvements were detected: need of a forgotten password functionality, and the fact that the face recognition image feedback was rather slow due to high CPU load. For the next phase, these two points will be improved. 

For Phase 4, another point to cover is the precise implementation of data gathering. 
For now, the data to collect is this:

\begin{verbatim}
{ "shapesid":"12142@gmail.com", 
"timestamp": “2013-10-21T13:28:06.419Z", 
"interaction_duration":"02:00:00", 
"activity_id":"4", 
"activity_name": “solitaire”, 
"emotion": “anger”, 
"status_code": “011”}
\end{verbatim}

Emotion data will not be used to adapt robot behavior in-situ, as during the hands-on it was observed not to be too reliable. The average of the highest confident emotion will be captured during each interaction and sent to SHAPES Data Lake for future analysis. 

\section{Conclusions}

The following paper describes the technical overview of the ARI robot prototype used in Pilot 1 of SHAPES pilot and the main hands-on training results gathered at Clinica Humana (Mallorca, \footnote{\url{https://www.clinicahumana.es/}}) between November-December 2021. Between February-March 2022, Phase 4 is taking place, where mentioned improvements will be tested with 5 targets users at Ca’n Granada and 1 caregiver for 2 weeks. Finally Phase 5, Deployment, will have the robot in complete autonomous mode with 10 service user participants and up to 2-3 carers: 1 month in individual homes and 5 months at the care-home. The robot will also be used in parallel in another pilot where it will deliver cognitive games \cite{dratsiou2022assistive}, which outputs will also be valuable for this pilot. 

\bibliography{main.bib}{}
\bibliographystyle{IEEEtran}
\end{document}